\pdfoutput=1
\documentclass[letterpaper, 10 pt, conference]{ieeeconf}  % Comment this line out if you need a4paper

\IEEEoverridecommandlockouts                              % This command is only needed if 
\usepackage[utf8]{inputenc}
\usepackage{graphicx} % for pdf, bitmapped graphics files
\usepackage{epigraph} % Epigraph quote in intro
\usepackage{multirow} % Multi-cellular tables.
\usepackage{amsmath, amsfonts} % Used for equation formatting
\usepackage{array} % Used for centred columns of fixed width.
\usepackage{subcaption} % Used for subfigures
\usepackage{tcolorbox} % For colours

\newcolumntype{P}[1]{>{\centering\arraybackslash}p{#1}} % Centred columns of fixed with - 'P{2cm}' rather than 'p{2cm}'

\definecolor{orange}{rgb}{1,0.5,0}

\renewcommand{\baselinestretch}{0.97}
\title{What Makes a Place? Building Bespoke Place Dependent Object Detectors for Robotics}

\author{
Jeffrey Hawke, Alex Bewley, Ingmar Posner% <- This % stops a space
\thanks{The authors are with the Oxford Robotics Institute, Department of Engineering Science, University of Oxford, Parks Road, Oxford OX1 3PJ, United Kingdom. Email: {\tt[jhawke, bewley, ingmar]@robots.ox.ac.uk}}
}

\begin{document}

\maketitle
\thispagestyle{empty}
\pagestyle{empty}

%!TEX root=main.tex

\begin{abstract}
This paper is about enabling robots to improve their perceptual performance through repeated use in their operating environment, creating local expert detectors fitted to the places through which a robot moves. We leverage the concept of `experiences' in visual perception for robotics, accounting for bias in the data a robot sees by fitting object detector models to a particular `place'. The key question we seek to answer in this paper is simply: how do we define a place? We build bespoke pedestrian detector models for autonomous driving, highlighting the necessary trade off between generalisation and model capacity as we vary the extent of the `place' we fit to. We demonstrate a sizeable performance gain over a current state-of-the-art detector when using computationally lightweight bespoke place-fitted detector models.
\end{abstract}

%!TEX root=main.tex

\section{Introduction}
\label{sec:intro}
%\epigraph{“We are all products of our environment; every person we meet, every new experience or adventure, every book we read, touches and changes us, making us the unique being we are.”} {\emph{C.J. Heck}}

Place matters in robotics. Mobile robots frequently traverse the same operating environment over and over, such as an autonomous car performing the same weekday commute, or an autonomous forklift moving goods around the same warehouse. In object detection for robotics, we care about achieving the best possible performance on the data the robot observes in its operating environment. We propose that relaxing the need for detector generalisation in favour of fitting to the operating environment enables better perception performance using a computationally simpler detector.

Prior work in robotics has demonstrated that it is possible to boost the performance of many vision-based systems by using place dependent models. Examples of these place dependent systems include object detectors \cite{hawke_wrong_2015, bewley2016imagenet}, terrain assessment \cite{berczi2016s}, and visual localisation systems \cite{linegar_made_2016,mcmanus_scene_2014}. A key question in considering all these environment specific models is simply: what defines an appropriate `place' for a model if we want to improve a perception system's performance? 

We focus on environment specific models for object detection, which we expect to see vary in performance with appearance change due to viewpoint, lighting, weather, season, and structural change in the environment.

Some of this performance variability will be spatially dependent, with the general scene structure and image texture remaining the same between images taken in the same location. In addition to this spatial variability, there are many temporal sources of environmental variation which will affect image appearance (and thus detector performance). This temporal variation includes appearance change from seasonal change, lighting change, and other dynamic objects operating in the same environment. In this paper we look at the influence of spatial variability on detector performance, and present a framework for improving performance through experience-based methods.

\begin{figure}[t]
	\centering
	\includegraphics[width=0.9\columnwidth]{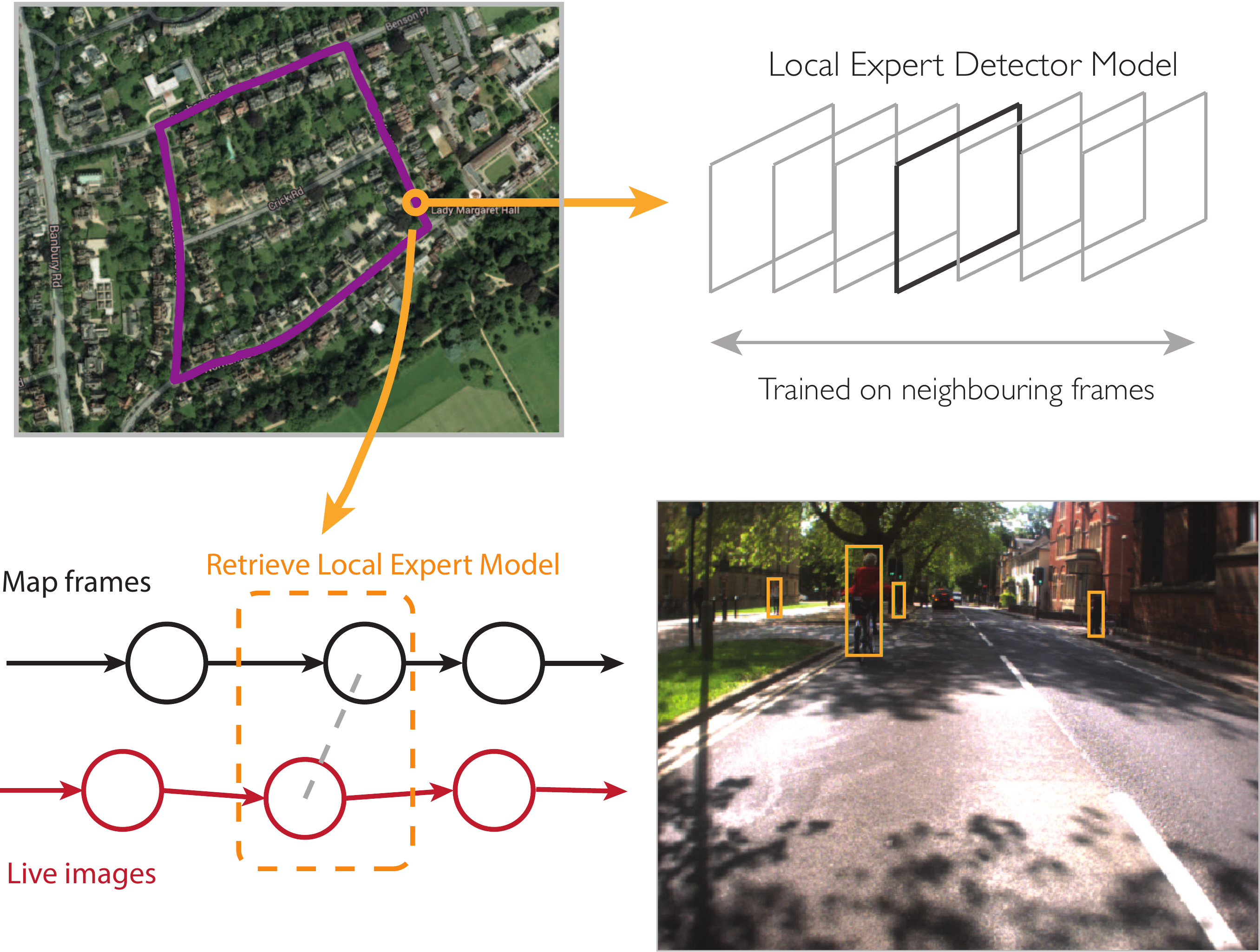}
	\caption{We can use the knowledge of where our robot is to improve object detection performance using a local expert model fitted to that specific place. Using lightweight detector models with limited model capacity we can boost performance beyond more computationally complex state-of-the-art detectors. We achieve this by limiting the variability of the negative (background) class our model is trained on, mitigating the model capacity limits of lightweight detectors.}
	\label{fig:intro-overview}
\end{figure} 

This paper is structured in two parts. We expect image data taken from the same location and time to have similar structure, enabling us to construct bespoke detectors on sequential images. Section \ref{sec:ideal-place-spatial} considers a scenario using this spatiotemporal definition of a place. We use an ensemble method: building a suite of local expert models for swathes of a robot's operating route using a nearest neighbour approach. These models are retrieved at run time using localisation. Section \ref{sec:ideal-place-similarity} extends this bespoke model methodology to use the nearest frames based on image appearance rather than spatial location. We approach all these experiments using pedestrians as the target object class as we have an interest in autonomous driving, though we expect the principles to hold across other object classes.

Our results demonstrate that overcoming limited model capacity (the ability of a classifier to separate the data classes) is the key benefit of using local expert detectors, enabling a lightweight detector to achieve a substantial performance increase over a more complex state-of-the-art pedestrian detector. We are able to do this by limiting the variability of the background (negative) class to local regions of the robot's operational envelope.

%!TEX root=main.tex

\section{Literature Review}
\label{sec:lit-review}
% Paragraph on pedestrian detection. Keep developments: better features, tend to use boosted decision trees.
% - Fast feature pyramids for object detection (Dollar PAMI14)
% - How far are we from solving pedestrian detection (Zhang/Benenson CVPR16)
% - Deeper look at pedestrians (Hosang/Benenson 2015 CVPR)
The approach we present in this paper can be applied to any object detection problem, but we focussed on pedestrian detection due to our interest in autonomous driving. Pedestrian detection has an extensive body of literature. Much of this work can be summarised as the development of progressively better engineered features \cite{dollar_fast_2014, zhang_informed_2014}, different classifiers \cite{wojek_performance_2008}, deformable parts models \cite{prioletti_part-based_2013}, or adding additional data or context \cite{enzweiler_efficient_2012,geronimo_2d3d-based_2010,walk_new_2010,ouyang_single-pedestrian_2013}.

Recent developments in deep learning in the broader computer vision community have lead to convolutional neural networks being applied to the problem \cite{hosang_taking_2015, yang_exploit_2016, cai2016unified}. These approaches are now achieving state-of-the-art performance but tend to be slower at run-time, which makes these ill-suited for use in resource constrained robotic applications.

% Relate this to mobile robotics applications. Use of additional information to boost performance in detection.
% - ROIs
% - Stixels, optical flow
% In addition to these developments in detection methods, various work has focussed on approaches which build on top of base detectors to boost performance. In object detection using monocular cameras this is can be accomplished by incorporating additional information such as optical flow \cite{walk_new_2010}, scene context \cite{ouyang_single-pedestrian_2013}, or stereo data \cite{enzweiler_efficient_2012} into the detector. This additional context helps guide the search through the image (such as region proposals \cite{geronimo_2d3d-based_2010}), reducing the false positive rate.

% Paragraph noting common failure cases: tends to have many more false positives, not false negatives.
Zhang \emph{et al.} \cite{zhang_how_2016} presented a comprehensive failure analysis of state-of-the-art pedestrian detectors on the Caltech pedestrian dataset \cite{dollar_fast_2014}. They hypothesised that many of the false negatives (i.e., cases where the detector fails to detect a pedestrian even though the pedestrian is large enough in the image) are due to dataset bias in training, which can be solved by augmenting the training dataset. Their analysis of common sources of false positives highlighted two categories of error: localisation errors where the false detection overlaps with a ground truth bounding box, and background errors where there is zero overlap with ground truth. These background errors were the most common error type, spuriously detecting parts of the background as pedestrians.

% Notion of experiences in robotics
% - Various MRG citations
% - Checkout my map - Gadd IROS16
%  EBC part 1:
% - Equivalent to HNM but self-supervised: NIPS
% - Performance boost by overfitting model to experience - similar class of data.
In previous work \cite{hawke_wrong_2015} we focussed on addressing these background errors in object detection by applying the concept of robot `experiences' from localisation and navigation \cite{churchill_experience-based_2013,linegar_work_2015,gadd_checkout_2016,paton_bridging_2016} to object detection. In an experience-based approach we accept that the world has many different forms of appearance and fit local experts to suit each experience. For perception tasks such as object detection, this means training our detector model to suit the class of data (e.g., season, spatial location) the robot currently experiences. We exploited the fact that mobile robots tend to be constrained to a particular operating environment to boost test performance by fitting models to data captured at a similar time of the year for the same traversed route. Notably, this experience-based detector outperformed a global model which has been trained on data across multiple experiences, highlighting the importance of defining the environment.

% Geospatial bias
% What makes paris look like paris - Doersch
Geospatial bias in imagery has been highlighted by Doersch et al. \cite{doersch_what_2012}, who showed that it is possible to automatically find distinctive visual elements for a certain geo-spatial area, comparing data between major cities (for example, Paris and London) and data between districts of a city (Paris). We also know that dataset bias \cite{torralba_unbiased_2011} is an influence on the performance of our object detectors, with the training data used being a systematic factor influencing performance across different test datasets \cite{benenson_ten_2014}. This data bias, however, provides a potential tool to be exploited. 

% TODO: Ensemble methods in machine learning:
% - Exemplar SVMs
% - Mixture of experts, boosting.
Malisiewicz et al. addressed bias in intra-class variability (such as different types of buses) in general purpose image recognition with Exemplar SVMs \cite{malisiewicz_ensemble_2011}, an ensemble approach which uses a nearest neighbour method to train a unique model for each positive training datum, fitted to a single positive sample and many thousands of negatives.

% Static scene EBC:
% Learning Scene-specific Pedestrian Detectors without Real Data - CVPR15
% Automatic Adaptation of a Generic Pedestrian Detector to a Specific Traffic Scene - CVPR11
% Transferring a Generic Pedestrian Detector Towards Specific Scenes - CVPR12
% Scene-Specific Pedestrian Detection for Static Video Surveillance - PAMI 13 -- COVERS ABOVE CVPR11/12
% Classifier Grids for Robust Adaptive Object Detection - CVPR09
Various approaches have been taken to leverage data bias to form scene specific classifiers, largely motivated by visual surveillance tasks. Hattori \emph{et al.} took the concept of scene specific classifiers to one extreme, training a unique detector model for a specific image location in a static scene using synthetic data \cite{hattori_learning_2015}. Wang \emph{et al.} used a transfer learning framework to adjust a generic pedestrian detector to a scene specific detector \cite{wang_scene-specific_2014}. They used visual cues to generate scene specific data in a self-supervised manner, weighting training data samples based on similarity to the observed test data using an affinity graph.

% Similar concept to EBC (but doesn't strictly consider experiences - trains models for GPS locations)
% Environment Adaptive Pedestrian Detection using In-vehicle Camera and GPS - VISAPP14
In the domain of mobile robotics, Suzuo \emph{et al.} divided a driven route into a fixed number of scenes using visual bag-of-words, then trained models using negative data gathered from each scene \cite{suzuo_environment_2014}. This has some similarity to our approach here; we build on the ideas from \cite{hawke_wrong_2015} and \cite{suzuo_environment_2014} to understand what influences an appropriate resolution for local expert detectors. Bewley and Upcroft presented similar evidence for the need to adapt object detectors to their operating environment in \cite{ROB:ROB21667}, augmenting a generalised object detector by validating each detection against a background model trained on environment specific data. This approach can also be viewed as exploiting place as context, with some similarities to methods which use context within an image to boost detector performance, e.g. \cite{chen_detection_2013}.

The idea of experience-based local expert models has since been applied to other perceptual tasks in robotics. Berczi \emph{et al.} \cite{berczi2016s} created bespoke terrain assessment classifier models for finite patches of a route driven with visual teach and repeat. These models detect when the path in front of a mobile robot has changed from the initial teach pass and previous repeat runs. They demonstrated a marked performance improvement over place-independent methods, comparing their bespoke place dependent models to a place independent learned classifier and a traditional heuristic based method.

Similarly, McManus \emph{et al.} \cite{mcmanus_scene_2014} and Linegar \emph{et al.} \cite{linegar_made_2016} both proposed using distinct scene specific patches to weakly localise across large appearance change.

These methods all have similar motivation: solving a simpler problem rather than a global one. However, why do these work so well? Why can't we simply use a generic model? In this paper we present evidence for \emph{why} place dependent models work well for solving these sorts of perception problems in robotics, and offer insight into how we should approach the construction of local expert models.

%!TEX root=main.tex

\section{Fitting to a Known Location and Time}
\label{sec:ideal-place-spatial}

We evaluate how to build bespoke place dependent models by firstly adopting a fine-grained, strict spatiotemporal definition of a place, where we construct a place dependent model for every single frame in a known route. We then immediately repeat the route, localise each image to our mapping run, then retrieve and apply the detector model associated with the localised map frame.

This is the best possible scenario we could hope for when fitting models to similar data. By using a strict spatiotemporal prior we are able to control most sources of variation in scene content (as the images are from the same spatial location) and appearance (as the images are from effectively the same time). In addition, we use a very fine grained approach, associating a model with every frame, rather than key frames. 

In practice, this will form the upper bound in performance for place dependent methods with this classifier and feature type, though we expect to be able to get close to this using experience-based visual localisation \cite{churchill_experience-based_2013} with models associated with mapping key frames.

\subsection{Experimental Approach}
%The best case scenario is evaluated using test data gathered from the same location and time as our training data. 

We select an urban driving dataset consisting of two consecutive loops of the same route in Oxford \cite{RobotCarDatasetIJRR}. Each loop is 1km in length and we fully annotated these with pedestrian labels. For each image (or model index frame), we fit a model to the closest $N$ images in time (including the frame itself), illustrated in Figure \ref{fig:intro-overview}. At run time, we localise each image to the alternate lap and retrieve that model, using the models built on the second lap during the first and vice versa.

\subsection{Object Detector and Training}
\label{sec:obj-detector}

Our object detector is a lightweight sliding window implementation using Aggregate Channel Features (ACF) on LUV images \cite{dollar_fast_2014}, with a linear Support Vector Machine (SVM) as the primary classifier. This is very similar to the original work in \cite{dollar_fast_2014}, except we use our own implementation of ACF with LibLinear \cite{fan_liblinear:_2008}, rather than the original boosted decision trees. This is a lightweight general purpose object detector, able to run at frame rate (20Hz) using a reasonable selection of scales. The ACF feature descriptor consists of ten channnels based on colour and gradient, making it suitable for detection tasks such as traffic lights \cite{philipsen2015traffic} and road signs \cite{mogelmose2015detection} in addition to the original purpose of pedestrian detection.

 We fit detector models to the nearest $N$ frames in time by performing hard negative mining (HNM) on each image using ground truth labels. This process runs a candidate detector model on the training images, adding the highest scoring false positives to the training data, retraining and iterating until convergence (or a maximum of $20$ iterations). The model is initially trained using the INRIA pedestrian dataset \cite{dalal_histograms_2005} for positive samples and an initial seed negative set (10 random patches from each INRIA negative image). We heavily penalise misclassification of training data by setting the SVM optimization parameter to a very large number ($C=100$), forcing the model to strongly fit to the training data presented. The training data is weighted based on the number of samples to ensure class balance.

%\begin{figure}
%\centering
%\includegraphics[width=\columnwidth]{images/ideal-place-spatial/ideal_place_expt_diagram.png}
%\caption{Fitting to a known location and time: We fit construct models associated with each frame in two laps of an annotated test route. Each model is trained on a sequence of frames centred around the model key frame.}
%\label{fig:ideal-place-diagram}
%\end{figure}

\subsection{Results}
To assess the merits of a precise spatiotemporal definition of a place, we compare our place-fitted detector to a generic model for our pedestrian detector trained only on a canonical dataset. This generic model is trained on the same initial training data as the place dependent models, but with HNM performed on the INRIA negative training data rather than data from the robot's environment. We refer to this baseline trained on generic data as the SVM+ACF model.

Table \ref{tab:ideal-place-results} outlines the aggregate performance metrics on our test dataset. We use three common metrics for comparing detector models: average precision, log-average miss rate \cite{dollar_fast_2014}, and maximum F1 score.

\begin{table}
    \vspace{2mm}
    \begin{tabular}{| P{3.1cm} | P{1.1cm} | P{1.5cm} | P{1.1cm} |}
        \hline
        Detector Model & Average Precision  & Log-Average Miss Rate & Maximum F1 Score \\ 
        \hline
        Baseline: SVM+ACF & 0.483 & 0.485 & 0.473 \\
        Baseline: FPDW \cite{dollar_fast_2014} & 0.481 & 0.440 & 0.491 \\
        Baseline: MSCNN \cite{cai2016unified} & 0.588 & 0.371 & 0.579 \\
        \hline
        P.D. Model: N=1  & 0.622 & 0.354 & 0.592 \\
        P.D. Model: N=10  & \textbf{0.689} & 0.292 & 0.651 \\
        P.D. Model: N=100  & \textbf{0.689} & \textbf{0.288} & \textbf{0.672} \\
        P.D. Model: N=1000  & 0.450 & 0.564 & 0.521 \\
        P.D. Model: N=$\sim$4100 (full lap) & 0.305 & 0.712 & 0.405 \\
        \hline
    \end{tabular}
    \caption{Performance metrics comparing our place dependent (P.D.) detector models to a number of generic baselines, where $N$ is the length of image sequence used to train a model. Higher is better for Average Precision and F1 Scores, whereas lower is better for Log-Average Miss Rate.}
    \label{tab:ideal-place-results}
\end{table}

\subsubsection{Benchmarks}
Results from two publicly available detectors are included to provide a baseline from which to assess potential performance gains. These are FPDW \cite{dollar_fast_2014}, and MSCNN \cite{cai2016unified}. We use models provided by the respective authors, trained on the Caltech pedestrian dataset \cite{dollar_fast_2014}. We acknowledge that there will be some bias in performance due to the training dataset, as INRIA has been shown to perform well across a diverse range of benchmark datasets \cite{zhang_how_2016}. Furthermore, as INRIA is a cropped dataset, it is only applicable to sliding window approaches making it unsuitable for training detectors like MSCNN.

FPDW is very similar to our SVM+ACF generic model, using ACF features but with a boosted decision tree classifier (AdaBoost) rather than an SVM. These two detectors are largely equivalent in performance and speed, able to run at frame rate (20Hz) for a reasonable range of scales on our Bumblebee2 camera using only moderate computational resources. MSCNN, a multi-scale convolutional neural network detector, is currently one of the best performing object detectors, but requires significant GPU based computational power and is only able to run at approximately 2Hz on a top end NVIDIA GTX graphics card. The authors state a run-time performance of 0.4s per frame on KITTI data (1242x375). Many robotic applications require a lower computational budget or higher frame rate.

\subsubsection{Detector Performance}
\begin{figure*}[htbp]
    \vspace{2mm}
    \begin{subfigure}[b]{0.98\textwidth}
        \centering
        \includegraphics[width=\textwidth]{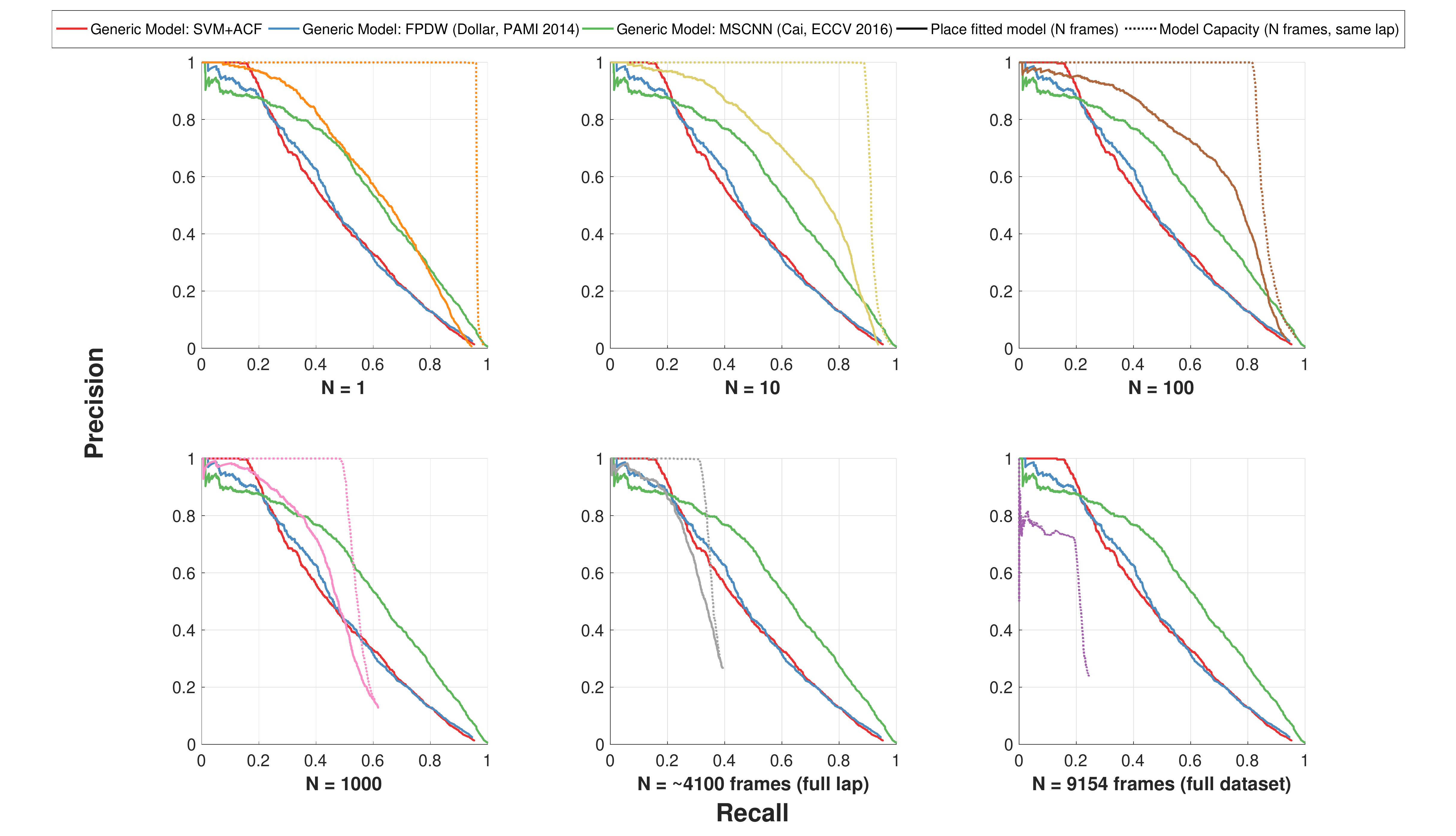}
        \caption{Precision - Recall (higher is better)}
    \end{subfigure}
    \begin{subfigure}[b]{0.98\textwidth}
        \centering
        \includegraphics[width=\textwidth]{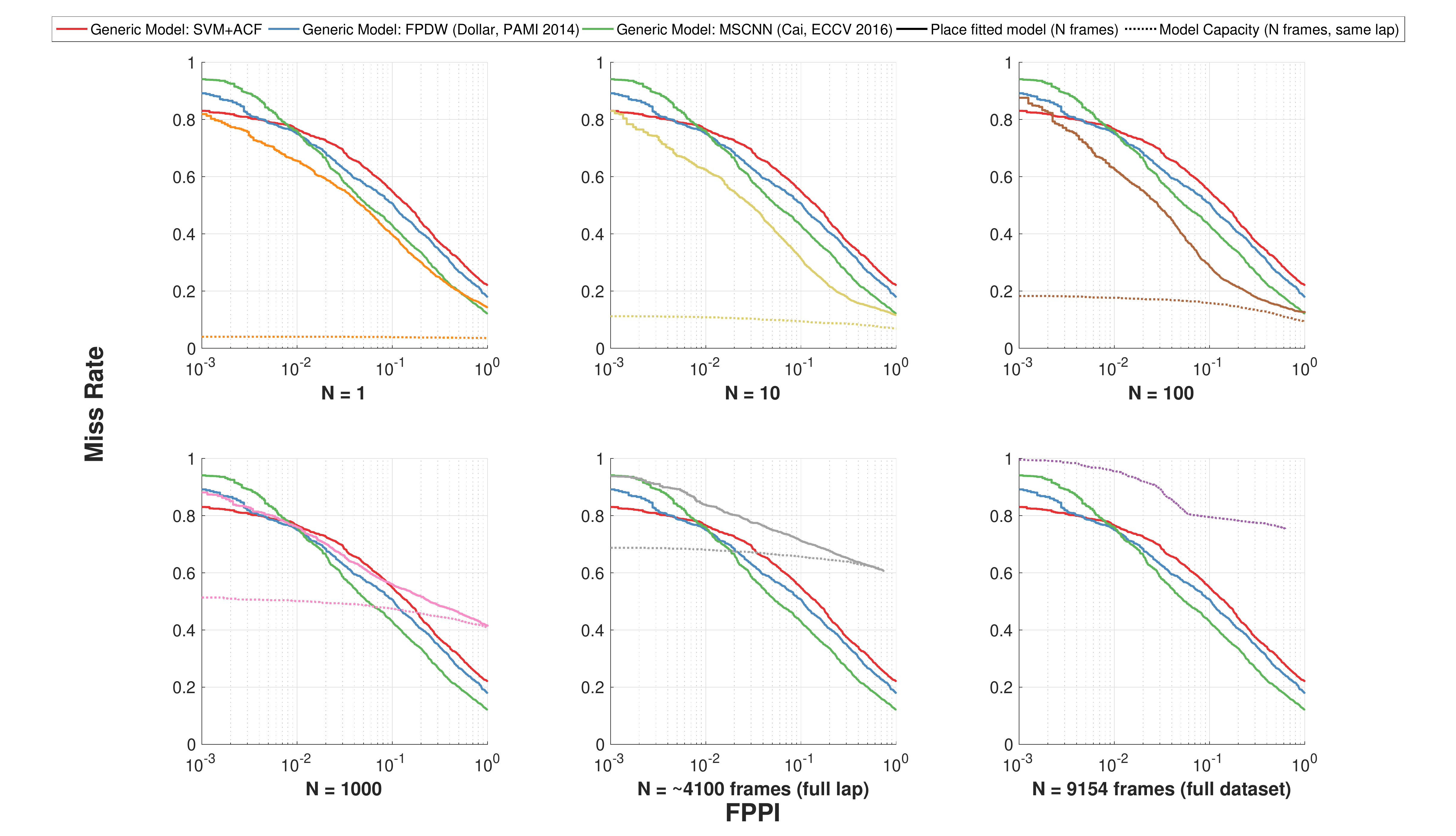}
        \caption{Miss Rate - False Positives Per Image (lower is better)}
    \end{subfigure}
    \caption{Performance of an ideal case place dependent detector compared to detectors using generic models. The place fitted detector uses models fitted on a frame-by-frame basis to a sequence of $N$ images centred on the closest matching frame from a second lap (the solid coloured lines). The dotted line on each plot represents the model capacity of the place fitted models, i.e., the performance when applying models to the same index frame they were trained on. This highlights the necessary trade-off between the size of negative class we fit to and the point at which we need the model to generalise.}
    \label{fig:ideal-place-spatial-pr-mrfppi}
\end{figure*}

The local expert detector outperforms all the baseline generic models (including the current state-of-the-art), with a twenty percentage point improvement over the reference model of the same type (SVM+ACF). Figure \ref{fig:ideal-place-spatial-pr-mrfppi} shows the characteristic performance curves for these models as the training sequence length varies. The dotted lines on each plot give us an appreciation for the model capacity, applying the model trained on each image to itself. Model capacity refers to be ability of a classifier to fit (separate) the data correctly without error, which will be determined by the classifier type, feature type, and the classes of data.

\subsubsection{Training Swathe Size}
Firstly, we note that using an extremely tightly fitted model (spatially, where the model is fitted to the single closest frame) offers a good improvement, but this is outperformed by models which are fitted to a slightly longer swathe ($N = 10-100$). There are a number of factors which will affect this. 

\begin{figure}[htbp]
    % \begin{subfigure}[t]{\columnwidth}
    % 	\centering
    % 	\includegraphics[width=\columnwidth]{images/ideal-place-spatial/localised_image_poses.eps}
    % 	\caption{Localised images will not necessarily align due to differences in the mapping and localisation trajectories. This variation makes fitting models to a single frame undesirable.}
    % 	\label{fig:ideal-place-spatial-loc-diagram}
    % \end{subfigure}
    % \begin{subfigure}[b]{\columnwidth}
    	\centering
    	\includegraphics[width=0.8\columnwidth]{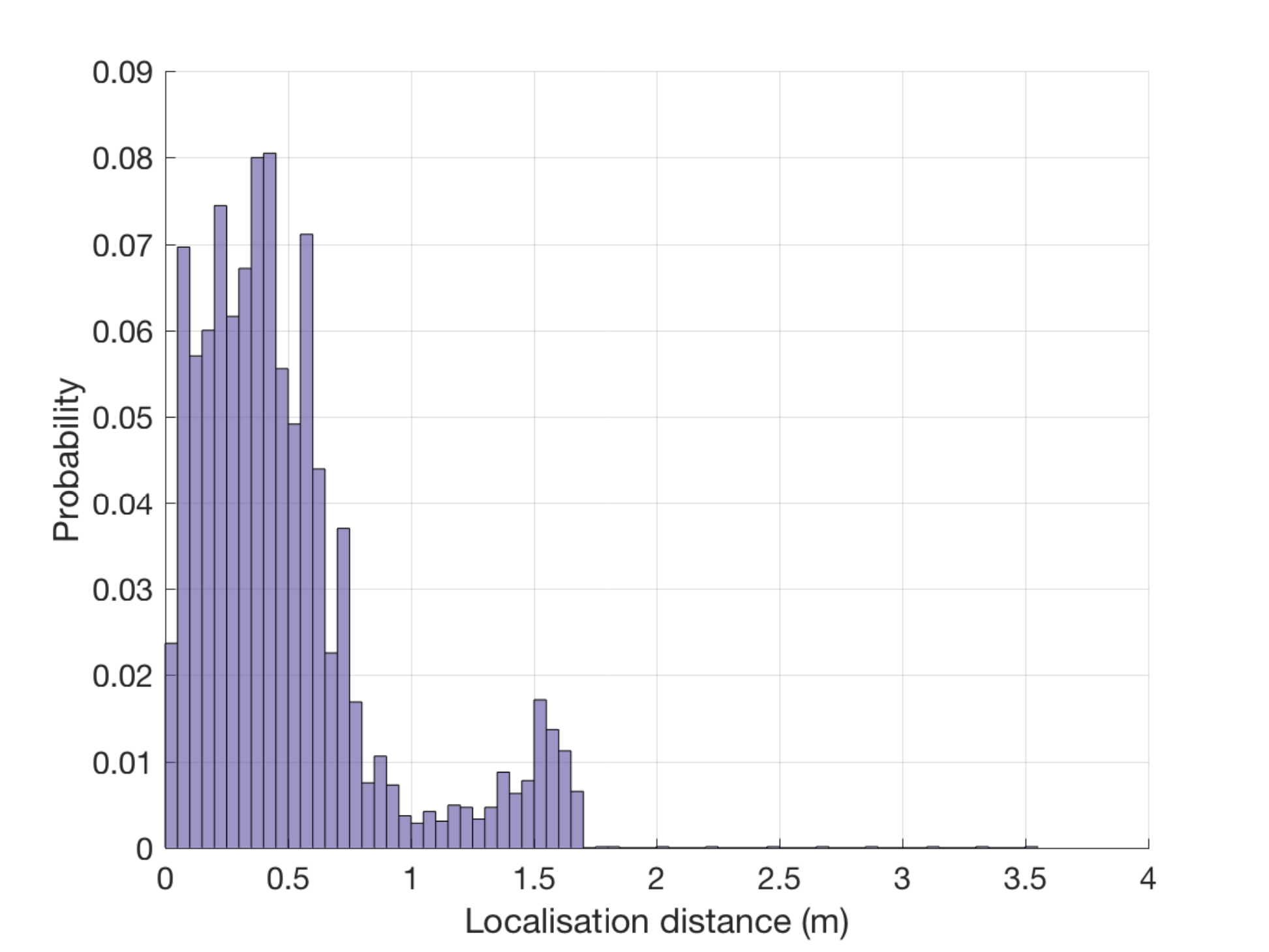}
    % 	\caption{In our best-case scenario fitting models to sequences of images, we see a median distance in pose of 0.40m between the map and localised model frames.}
    % 	\label{fig:ideal-place-spatial-loc-hist}
    % \end{subfigure}
    \caption{Variation in pose between localised images on subsequent laps (a median of 0.4m for our dataset) will limit the efficacy of models fitted to a single frame.}
    \label{fig:ideal-place-spatial-loc-imgs}
\end{figure}

Firstly, longer training swathes will inevitably add more training data, which will generally help performance. In addition, these longer swathes will also help to account for the differences in pose between the image and the model index frame. This pose error is small (as per Figure \ref{fig:ideal-place-spatial-loc-imgs}), but will limit the efficacy of models fitted to a single frame due to variation in perspective and scene content.

On the other hand, using a large sequence of frames comes with a penalty. As we are forcing the model to fit the training data firmly (using a high value for the SVM kernel parameter), we begin to see a drop in performance due to limited model capacity once the negative set is sufficiently large such that we lose class separability.

\begin{figure}[htbp]
    \vspace{2mm}
	\centering
	\includegraphics[width=0.8\columnwidth]{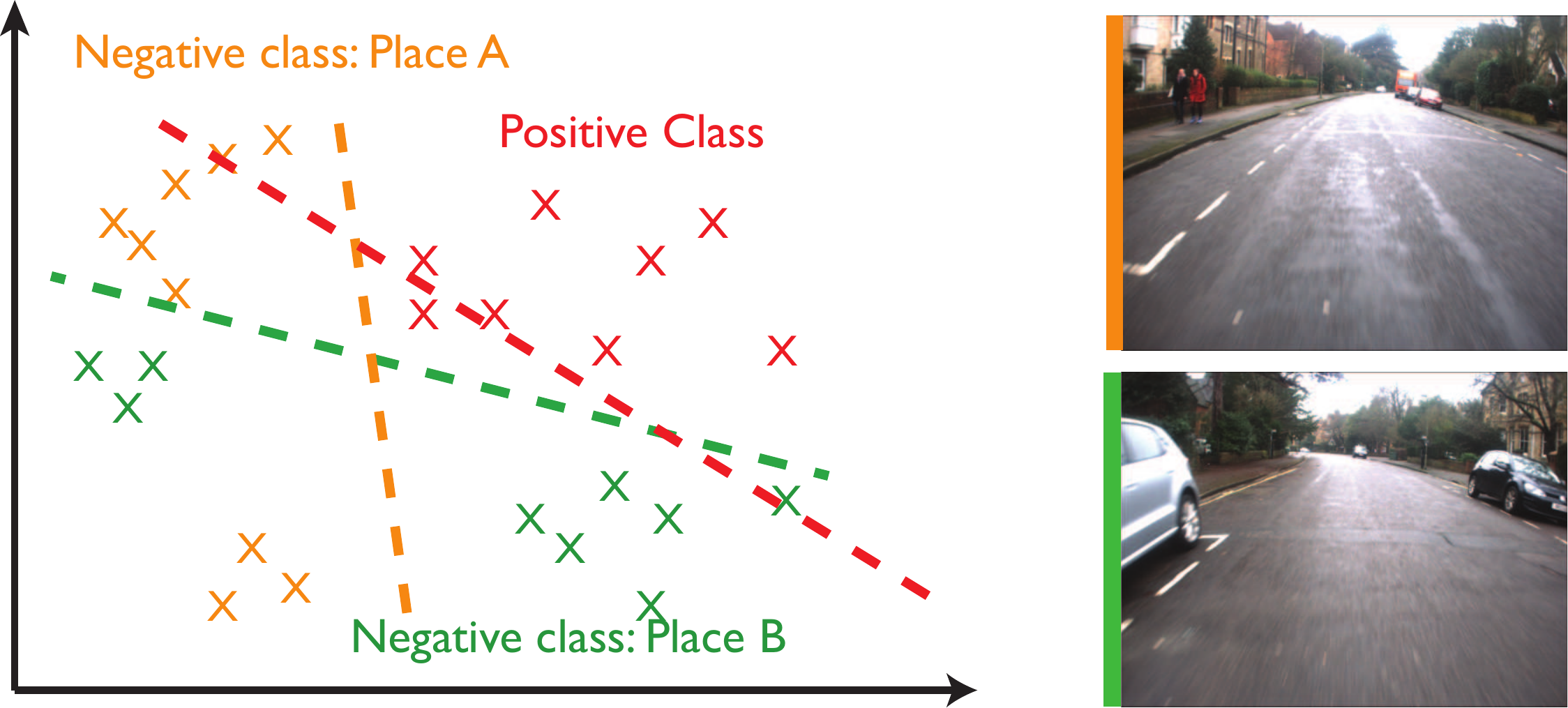}
	\caption{This illustration outlines a simple example of where using a simpler (faster) classifier cannot separate the positive and negative class perfectly due to finite model capacity, indicated by the red dashed decision boundary. Reducing the background class boosts performance by training models for a single environment (e.g., the orange or green images), giving full class separability with a collection of models.}
	\label{fig:model-cap}
\end{figure}

We start to see the impact of finite model capacity on performance at a swathe size of between 100-1000 frames. At 1000 frames we have lost class separability, in which the linear SVM classifier is unable to separate the positive and negative training data with no errors (illustrated conceptually in Figure \ref{fig:model-cap}). As we weight the training data for class balance, this lack of class separability results in the hard negative mining reaching the maximum number of cycles rather than converging. 

This performance drop due to model capacity is even clearer in the \emph{global model} case, where we fit the model to the entire dataset. In this case, we only have a model capacity plot as we train the model on every image ($N = 9154$). Even though the global model is trained on the exact test data, it is unable to separate the positive and negative class, requiring the model to generalise. 

\subsubsection{Concluding remarks}
With lightweight pedestrian detectors performance is boosted by using local expert models which have a very specific place of operation. The extent of a place — measured by a fixed swathe size — should contain multiple frames to avoid sensitivity to the pose difference between training and testing data. Conversely, a narrow definition of a place with minimal variation in the background class is desirable to enhance class separability with finite model capacity. In our experiments, models trained with sequences of $10-100$ frames showed the most significant improvement.

\begin{figure*}[htbp]
	\centering
	\includegraphics[width=\textwidth]{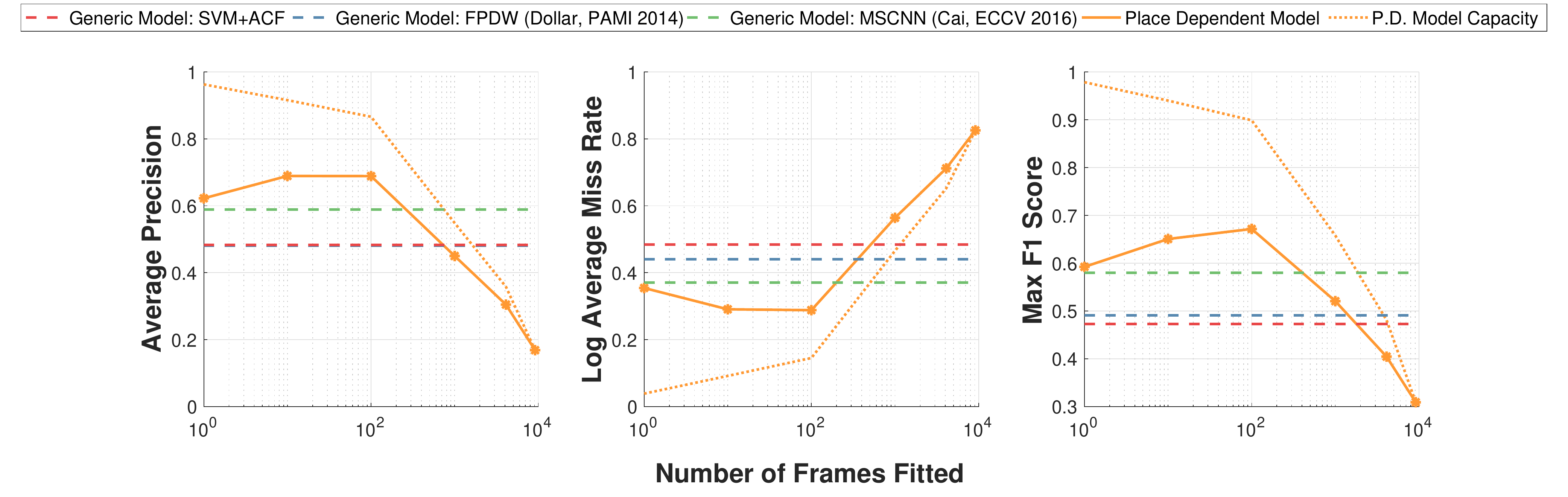}
	\caption{Aggregate performance metrics for place dependent detector models as a function of $N$, the number of frames in the nearest neighbour training swathe. The dotted line represents the model capacity. Note that our generic model (SVM+ACF) and FPDW score almost identically in average precision.}
	\label{fig:ideal-place-spatial-metrics}
\end{figure*}

%\input{04_generalised_place.tex}

%!TEX root=main.tex

\section{Linking Appearance to Place Dependent Models}
\label{sec:ideal-place-similarity}

% Motivational speech bit
Spatial location is important, but we also expect to see image appearance change over time: seasons, lighting, and other moving objects. From localisation literature we know that image appearance change is significant, and has a dramatic effect on the performance of perception systems \cite{churchill_experience-based_2013}.

\begin{figure}[htbp]
    \begin{subfigure}[b] {0.32\columnwidth}
        \centering
        \includegraphics[width=\columnwidth]{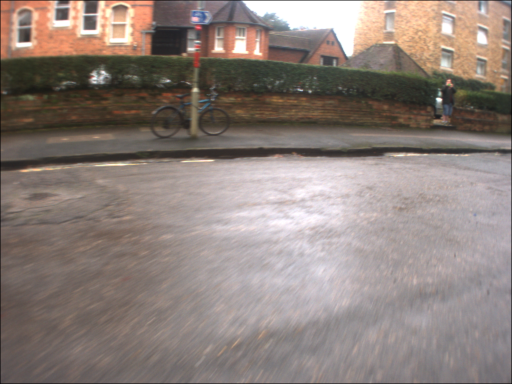}
        \caption{}
        \label{fig:nthoxfjan_a}
    \end{subfigure}
    \begin{subfigure}[b]{0.32\columnwidth}
        \centering
        \includegraphics[width=\columnwidth]{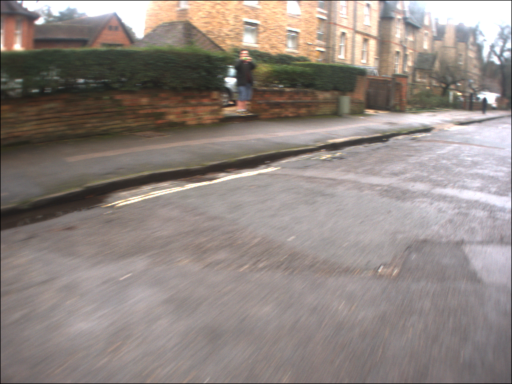}
        \caption{}
        \label{fig:nthoxfjan_b}
    \end{subfigure}
    \begin{subfigure}[b]{0.32\columnwidth}
        \centering
        \includegraphics[width=\columnwidth]{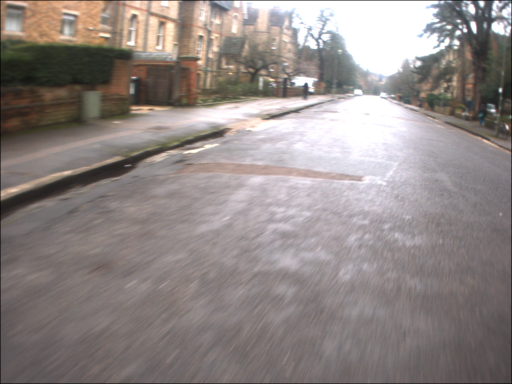}
        \caption{}
        \label{fig:nthoxfjan_c}
    \end{subfigure}
    \begin{subfigure}[b]{\columnwidth}
        \centering
        \includegraphics[width=\columnwidth]{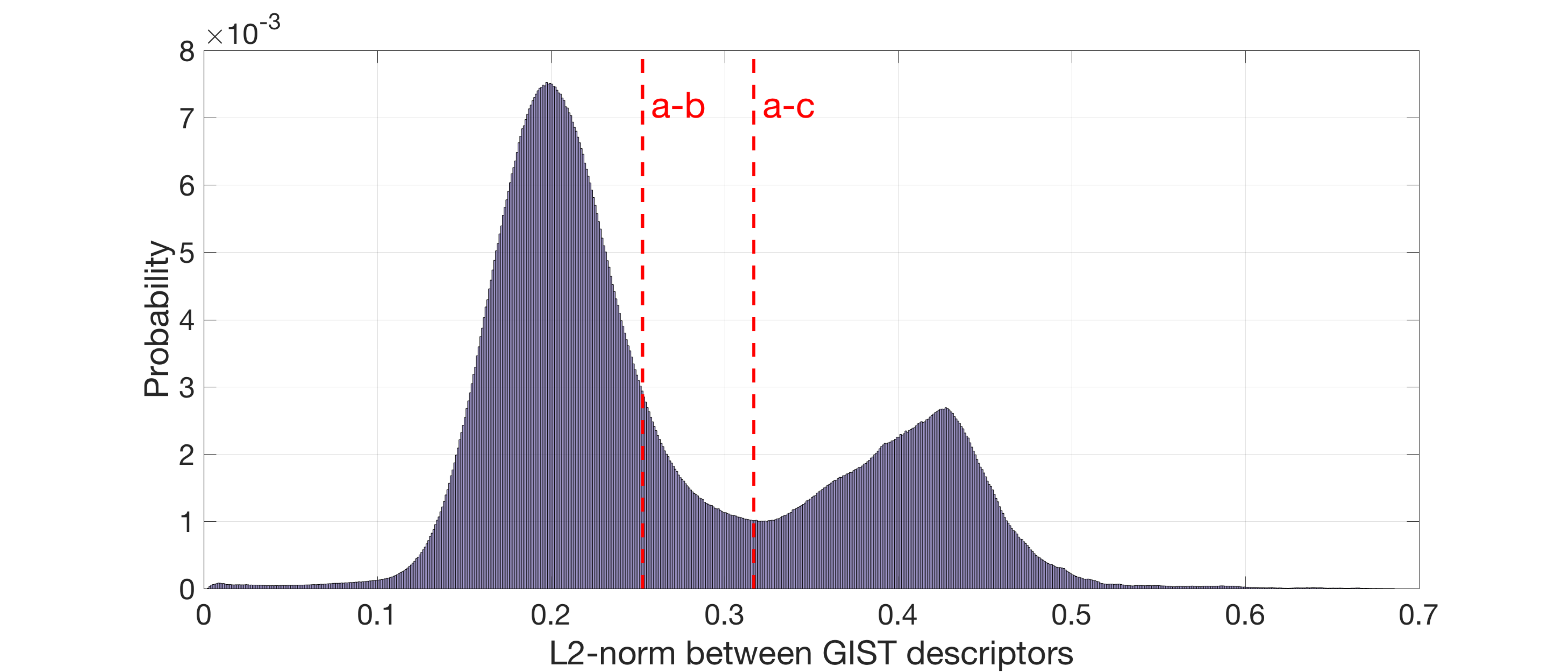}
        \caption{}
        \label{fig:nthoxfjan_gist_dist_hist}
    \end{subfigure}
    
    \caption{Images \ref{fig:nthoxfjan_a} through \ref{fig:nthoxfjan_c} are taken one second apart (approximately 3 metres), but differ substantially when comparing the GIST descriptors. This is due to the rapidly changing perspective rather than any substantive change in scene content. The histogram in \ref{fig:nthoxfjan_gist_dist_hist} shows the similarity distribution for all image pairs in this route, with the red lines showing the similarity between image pairs \ref{fig:nthoxfjan_a}-\ref{fig:nthoxfjan_b} and  \ref{fig:nthoxfjan_a}-\ref{fig:nthoxfjan_c}. These fall in the 61st and 71st percentile, highlighting the difference in similarity between spatially close images.}
    \label{fig:img-appearance-change}
\end{figure}

\begin{figure} [htbp]
    \begin{subfigure}[b]{\columnwidth}
        \centering
        \includegraphics[width=0.95\columnwidth]{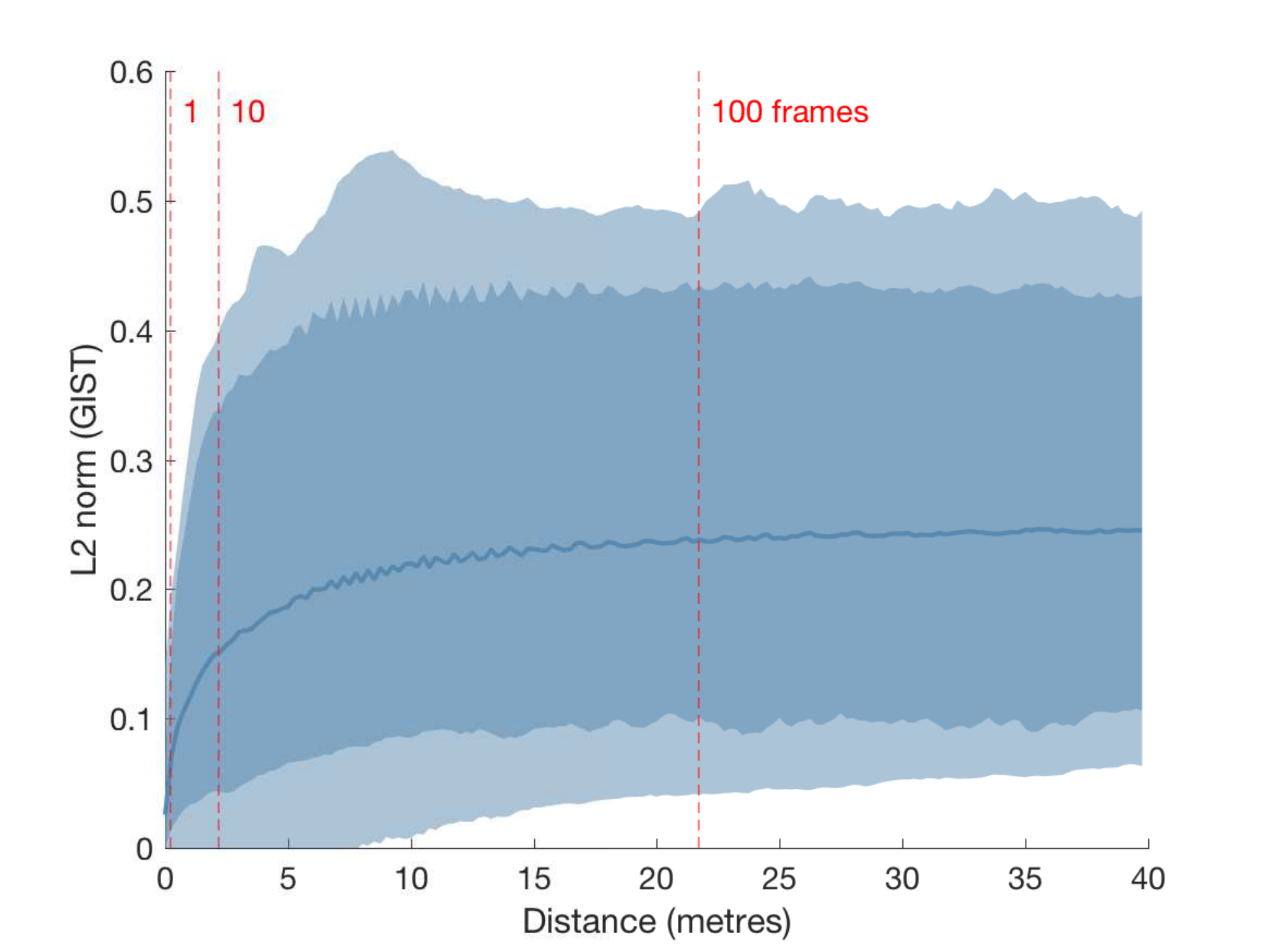}
        \caption{$l^2$-norm between GIST}
        \label{fig:gist-path}
    \end{subfigure}
    \begin{subfigure}[b]{\columnwidth}
    	\centering
    	\includegraphics[width=0.95\columnwidth]{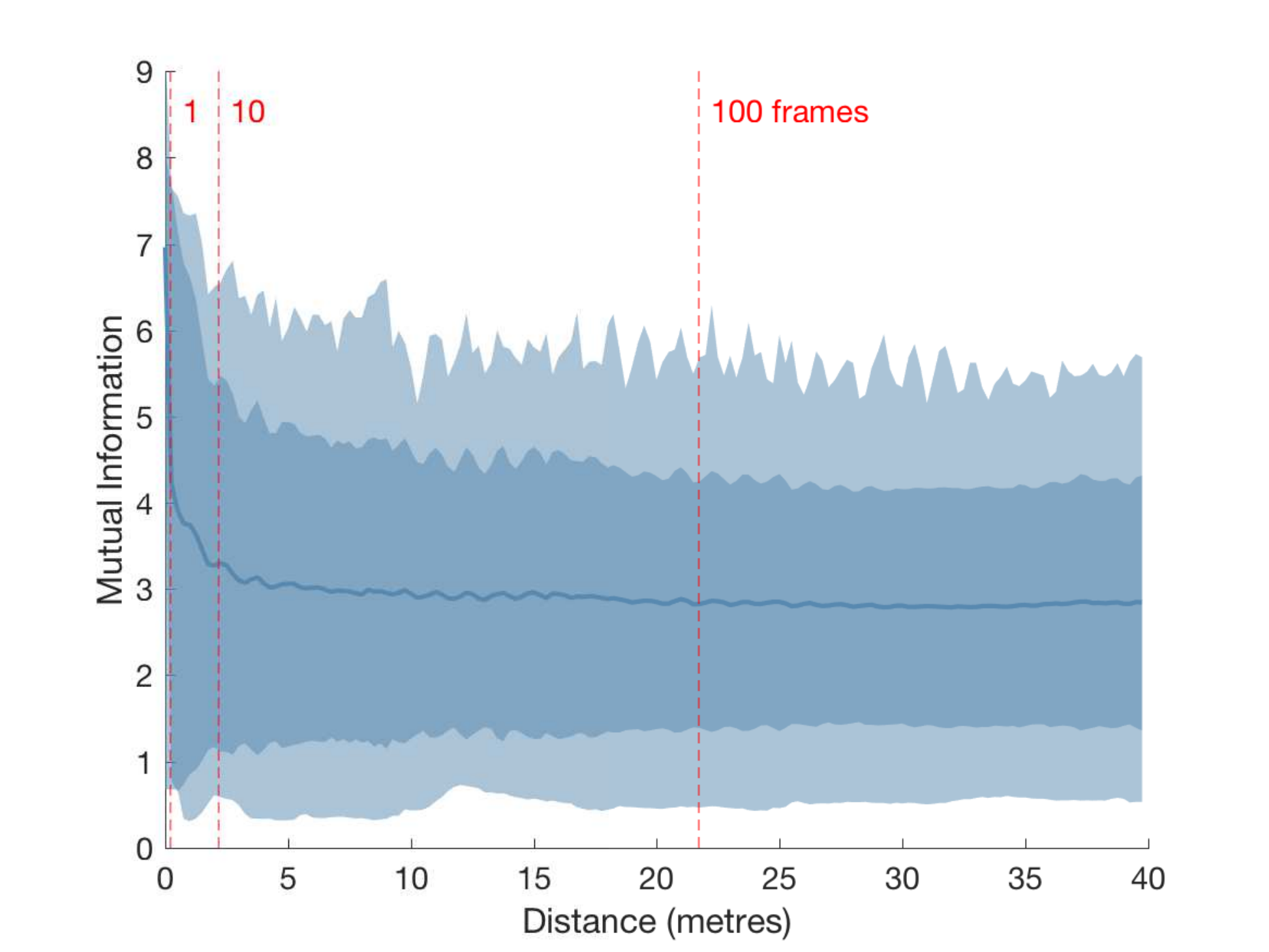}
    	\caption{Mutual Information}
    	\label{fig:mi-path}
    \end{subfigure}
\caption{Image similarity metrics correlate with the distance travelled along the robot's trajectory. The mean similarity is shown by the solid line, and the filled polygons represent the $2\sigma$ and maximum extent. The red dashed lines indicate the distance travelled within $N=1,10,100$ frames during this test route at the vehicle's average speed of $4.3$ ms$^{-1}$.}
\label{fig:sim-path}
\end{figure}

As we focus on boosting performance by reducing the extent of the background class our model fits to, it logically holds that we should build models on images with similar appearance. In many cases this is influenced by spatial proximity, as higher similarity is correlated with spatial proximity (shown in Figure \ref{fig:sim-path}). The previous section (Section \ref{sec:ideal-place-spatial}) addresses this. However spatial correlation is not guaranteed, as appearance metrics vary with changes in scene content and perspective. A key example of this is a sequence of images taken driving around a corner: these will be dissimilar by appearance metrics but are spatially close (Figure \ref{fig:img-appearance-change}).

\subsection{Experimental Approach}
% Experimental approach
We adopt the same experimental scenario as in Section \ref{sec:ideal-place-spatial}, where we have two consecutive laps of an urban driving route with ground truth localisation and pedestrian annotations. For every image, we localise to the closest frame in the alternate lap, retrieving the detector model associated with the map frame and apply it. The object detector setup and training is as per Section \ref{sec:obj-detector}.

However, to build the detector models we now use the $N$ nearest frames based on image similarity metrics, rather than $N$ nearest temporal frames. We use two similarity measures: the $l^2$-norm between two GIST global image descriptors \cite{douze_evaluation_2009}, and mutual information \cite{viola1997alignment} between two intensity images. 
 
Our results in Section \ref{sec:ideal-place-spatial} indicate that our detector model was able to fit well for a place definition of between $1-100$ frames, with $1000$ frame detector models suffering in performance due to model capacity (Figure \ref{fig:ideal-place-spatial-metrics}). This result agrees with the data we see in Figure \ref{fig:sim-path}, which indicates that the correlation between image appearance and distance no longer applies beyond $100$ frames at the average vehicle speed. We focussed our experiments on models built using the same number of frames as before ($N=1-100$), starting from a single frame and increasing by a decade at each step. 

\subsection{Results}
% Results and discussion
To assess this approach, we use the same experimental methodology as in Section \ref{sec:ideal-place-spatial}, using localisation to retrieve the model associated with the map frame closest to the current image.
 
\begin{figure*}[htbp]
 \centering
 \includegraphics[width=\textwidth]{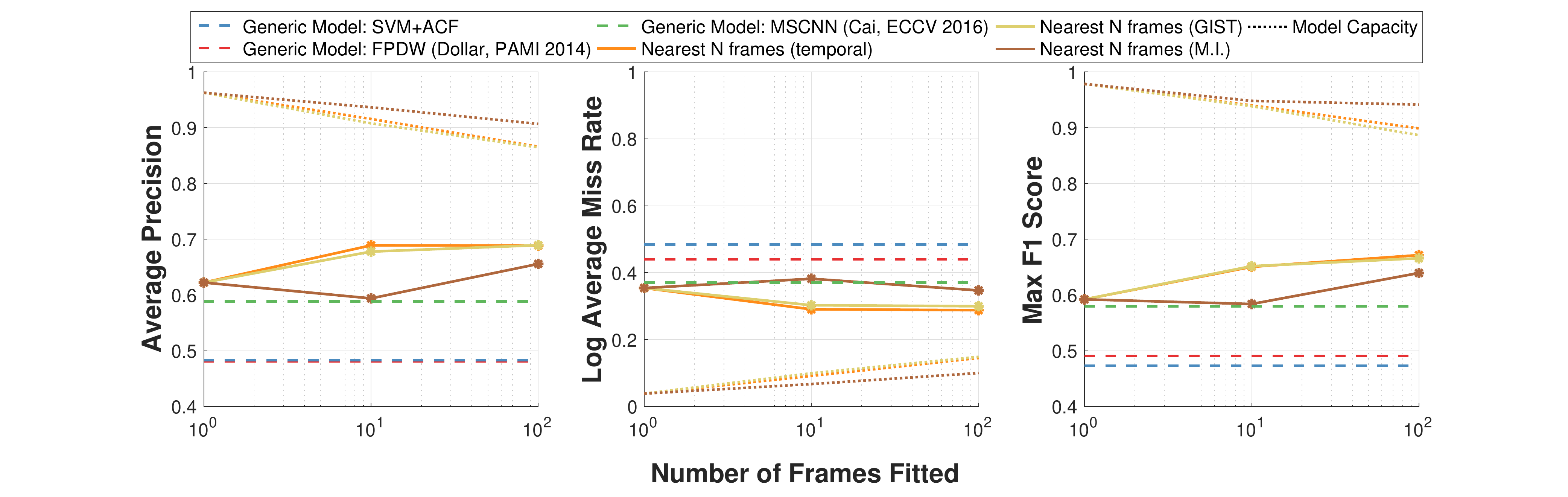}
 \caption{Aggregate performance metrics for models fitted to the nearest $N$ frames using image similarity and sequences. The spatiotemporal and GIST based models perform similarly, surpassing the baseline generic models. Two baseline models (SVM+ACF, FPDW) score identically in average precision, and the dotted lines indicate the SVM's limited model capacity.}
 \label{fig:ideal-place-similarity-metrics}	
 \end{figure*}

In Figure \ref{fig:ideal-place-similarity-metrics} and Table \ref{tab:ideal-place-similarity-results} we see a performance boost over the reference baseline detectors. Using the nearest images based on GIST yields more or less equivalent performance. 

\begin{figure} [tbp]
    \begin{subfigure}[b]{0.49\columnwidth}
        \centering
        \includegraphics[width=\columnwidth]{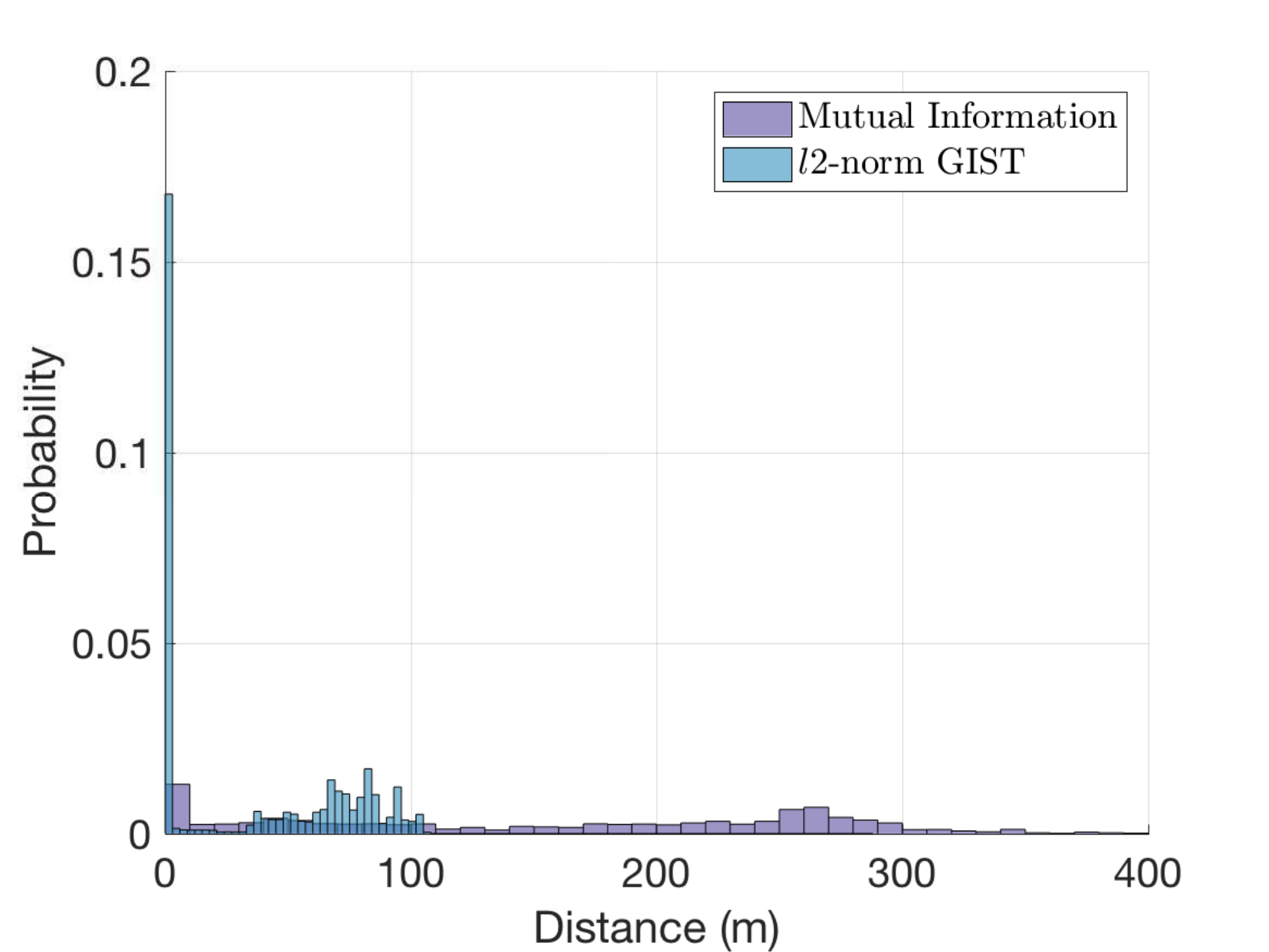}
        \caption{10 frame models}
        \label{fig:similarity-10frames}
    \end{subfigure}
    \begin{subfigure}[b]{0.49\columnwidth}
    	\centering
    	\includegraphics[width=\columnwidth]{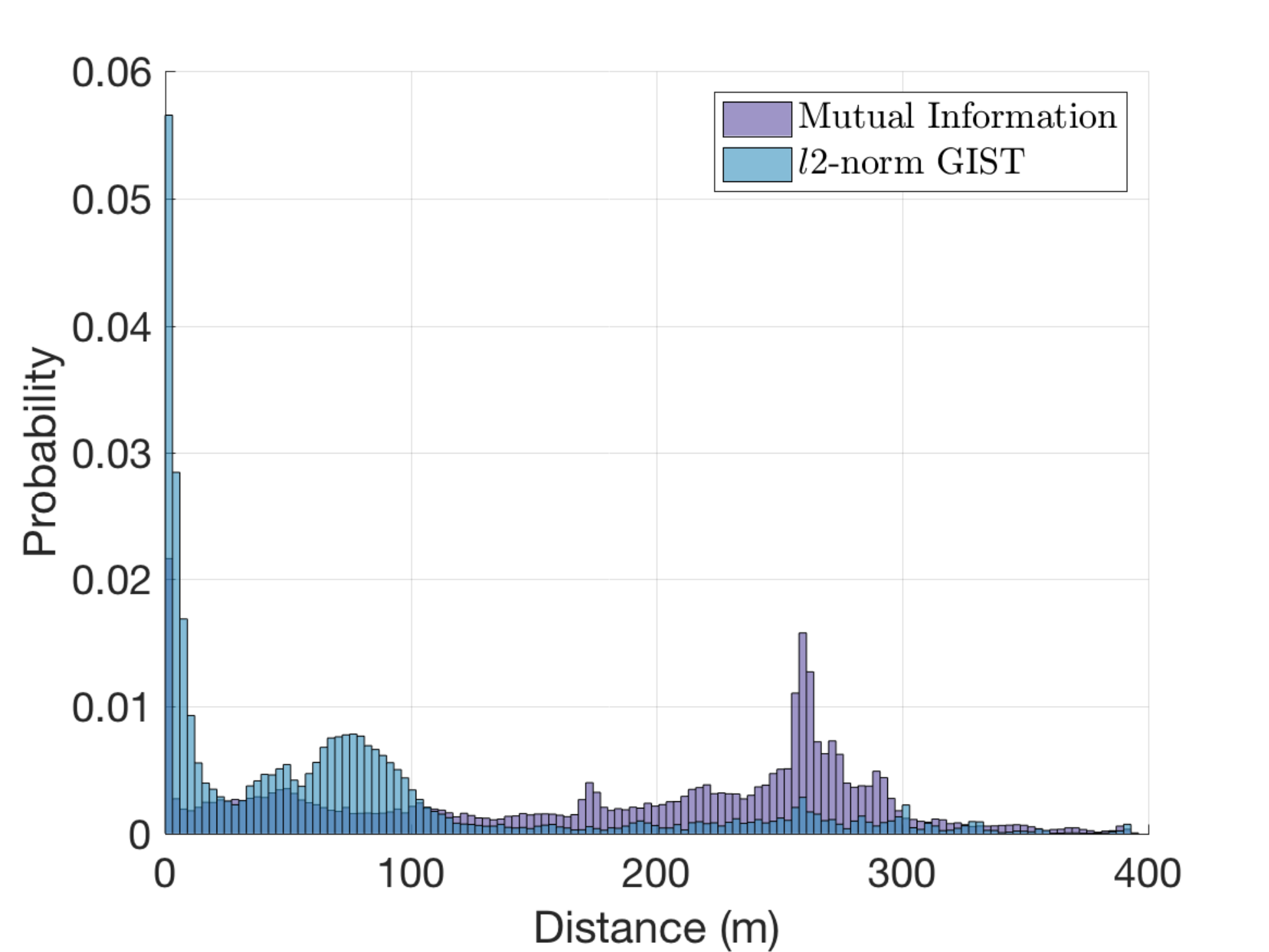}
    	\caption{100 frame models}
    	\label{fig:similarity-100frames}
    \end{subfigure}
\caption{These histograms highlight the spatial distance in models' training data when using different image similarity metrics. For models built using the nearest $N$ images (spatially), a $100$ frame model will on average have been trained on data up to $22$ metres from the index frame (as in Figure \ref{fig:sim-path}). In contrast, when using the nearest $100$ frames by either image similarity metric this can be up to $400$ metres.}
\label{fig:similarity-histograms}
\end{figure}

This logically holds: for our dataset we expect the temporally adjacent frames to be very similar in appearance, with the exception of perspective change when turning corners. Using mutual information is less beneficial: this still helps boost performance, but is not as informative as grouping images based on time or GIST. In Figure \ref{fig:mi-path} we note that the correlation between mutual information and distance plateaus at $5$ metres, unlike with GIST. The mutual information models are also trained on data which is much further from the source frame than GIST (Figure \ref{fig:similarity-histograms}). We expect that mutual information between each colour channel rather than intensity would be better, though this is unlikely to improve significantly over the temporal sequence approach.
 
It is likely that we could use image appearance to fit better models for larger groups of data, but we are limited by model capacity before this effect is visible. With a higher capacity model (e.g., deep learning approaches or different kernels) we may be able to exploit image appearance to a greater degree where we lack robust localisation.

\begin{table}[tbp]
    \centering
    \begin{tabular}{| P{2.8cm} | P{1.3cm} | P{1.5cm} | P{1.3cm} |}
        \hline
        Model & Average Precision  & Log-Average Miss Rate & Maximum F1 Score \\ 
        \hline
        Baseline: SVM+ACF & 0.483 & 0.485 & 0.473 \\
        Baseline: FPDW \cite{dollar_fast_2014} & 0.481 & 0.440 & 0.491 \\
        Baseline: MSCNN \cite{cai2016unified} & 0.588 & 0.371 & 0.579 \\
        \hline
        P.D. Spatial: N=1  & 0.622 & 0.354 & 0.592 \\
        P.D. Spatial: N=10  & \textbf{0.689} & 0.292 & 0.651 \\
        P.D. Spatial: N=100  & \textbf{0.689} & \textbf{0.288} & \textbf{0.672} \\
        \hline
        P.D. $l^2$ GIST: N=10  & 0.678 & 0.303 & 0.652 \\
        P.D. $l^2$ GIST: N=100  & \textbf{0.689} & 0.300 & 0.666 \\
        \hline
        P.D. Mut. Inf.: N=10  & 0.594 & 0.382 & 0.584 \\
        P.D. Mut. Inf.: N=100  & 0.656 & 0.347 & 0.640 \\
        \hline
    \end{tabular}
    \caption{Aggregate performance metrics comparing our bespoke place dependent (P.D.) detector models to a number of generic baselines. A higher number is better for Average Precision and F1 Scores, lower is better for Log-Average Miss Rate. The $N=1$ case is the same for all methods as this model is trained on the closest localised frame.}
    \label{tab:ideal-place-similarity-results}
\end{table}

%\input{06_deep_learning.tex}

%!TEX root=main.tex

\section{Conclusions}
\label{sec:conc}

In robotics we have the opportunity to exploit knowledge of the world around our robot to improve performance of vision systems. In this paper we demonstrate how prior knowledge about what a robot will see (by repeated operation along the same route) can be used to boost performance beyond what a more complex state-of-the-art object detector can achieve. However, defining the place of operation is not trivial for mobile robots. We present an assessment of approaches to building bespoke place dependent detector models, investigating which factors influence performance. 

Our results indicate that we can heavily leverage a strong spatiotemporal prior, fitting models to a narrow window of operation. Relaxing the need for generality aids performance as we can ensure our models fit a smaller negative background class. This approach avoids the trade off between model capacity and generalisation. When using appearance to build models, we are able to achieve effectively the same performance on groups of similar images. 

As model capacity is a major influence on performance of these local models, we anticipate that using higher capacity classifiers (such as deep learning methods or more complex SVM kernels) will enable fitting to larger operating environments. Computational requirements will dictate the choice of classifier used in these local expert detectors. We also only considered expert models fitted to a fixed number of images. It is likely that larger visually uniform sections of a route could be adequately fitted with a single model, whereas more varied routes will need a more specialised model fitted to a shorter section.  

From a practical view, we see significant potential for this approach to be coupled to experience-based visual localisation systems \cite{churchill_experience-based_2013,linegar_work_2015,gadd_checkout_2016,paton_bridging_2016}, using their ability to localise to many parallel `experiences' of the same location. These bespoke detector models are sensitive to the data they are applied to: if the scene content changes substantially from the training environment, it is likely that the detector will generate false positive errors due to the change in background class. This brittleness could be addressed by using models constructed from appearance rather than spatiotemporal sequences, or by using the ability of experience-based localisation to localise across different appearances of the same place, with detector models associated with each experience of the same spatial location.

\section*{Acknowledgment}
\label{sec:ack}
The authors would like to acknowledge the support from the EPSRC Programme Grant DFR01420 as well as the Advanced Research Computing at the University of Oxford.

%\bibliographystyle{IEEEtran}
%\bibliography{iros-2017}

\end{document}